\documentclass[11pt,letterpaper]{amsart}

\usepackage[margin=1.25in]{geometry}
\usepackage[foot]{amsaddr}
\usepackage{times}
\usepackage{epsfig}
\usepackage{graphicx}
\usepackage{subcaption}
\usepackage{amsmath}
\usepackage{algorithm}
\usepackage{algorithmic}
\usepackage{amssymb}
\usepackage{adjustbox}
\usepackage{booktabs}
\usepackage{dblfloatfix}
\usepackage[shortlabels]{enumitem}
\usepackage{bm}
\newcommand{\R}{\mathbb{R}}
\newcommand{\cX}{\mathcal{X}}
\newcommand{\cE}{\mathcal{E}}

\DeclareMathOperator*{\argmax}{arg\,max}
\DeclareMathOperator*{\argmin}{arg\,min}
\usepackage{siunitx}
\usepackage{array}
\newcolumntype{L}[1]{>{\raggedright\let\newline\\\arraybackslash\hspace{0pt}}m{#1}}
\newcolumntype{C}[1]{>{\centering\let\newline\\\arraybackslash\hspace{0pt}}m{#1}}
\newcolumntype{R}[1]{>{\raggedleft\let\newline\\\arraybackslash\hspace{0pt}}m{#1}}
\usepackage{multirow, makecell}

\usepackage{hyperref}

\newtheorem{theorem}{Theorem}

\begin{document}

\title[The ProxLogBarrier adversarial attack]{A principled approach for generating adversarial images under non-smooth dissimilarity metrics}

\author{
Aram-Alexandre Pooladian$^1$
\and
Chris Finlay$^1$
\and
Tim Hoheisel$^1$
\and 
Adam Oberman$^{1}$
}

\address{$^1$ Department of mathematics and statistics, McGill University}
\email{\{aram-alexandre.pooladian,christopher.finlay\}@mail.mcgill.ca}
\email{tim.hoheisel@mcgill.ca}
\email{adam.oberman@mcgill.ca}
\thanks{AO supported by AFOSR grant FA9550-18-1-0167}

\maketitle

\begin{abstract}
Deep neural networks perform well on real world data but are prone to
adversarial perturbations: small changes in the input easily lead to
misclassification. In this work, we propose an attack methodology not only for
cases where the perturbations are measured by $\ell_p$ norms, but in fact any
adversarial dissimilarity metric with a closed proximal form. This includes, but
is not limited to, $\ell_1, \ell_2$, and $\ell_\infty$ perturbations; the $\ell_0$
counting ``norm" (i.e. true sparseness); and the total variation seminorm, which
is a (non-$\ell_p$) convolutional dissimilarity measuring local pixel changes.
Our approach is a natural extension of a recent adversarial attack method, and
eliminates the differentiability requirement of the metric. We demonstrate our
algorithm, ProxLogBarrier, on the MNIST, CIFAR10, and ImageNet-1k datasets. We
consider undefended and defended models, and show that our algorithm easily
transfers to various datasets. We observe that ProxLogBarrier outperforms a host
of modern adversarial attacks specialized for the $\ell_0$ case. Moreover, by
altering images in the total variation seminorm,  we shed light on a new class
of perturbations that exploit neighboring pixel information.
\end{abstract}
\section{Introduction}
Deep neural networks (DNNs) have strong classification abilities on training and validation datasets. 
However, they are vulnerable to adversarial images, which are formally defined
as imperceptibly small changes (in a given dissimilarity metric) to model input
that lead to misclassification \cite{origin_adversarial,FGSM}. This behavior
could mean several things: the model is overfitting on some level; the model is
under-regularized; or this is simply due to complex nonlinearities in the model. This has lead to several lines of work in the deep learning community: the generation of adversarial images, defending against these adversarial attacks, and lastly determining \textit{which} dissimilarity metric to consider.  

Regarding the latter, it is not obvious what ``imperceptibly small" means, and
recent work has demonstrated adversarial image generation beyond $\ell_p$ norms
by considering \textit{deformations} instead of perturbations \cite{adef}.
There is also the problem of generating ``realistic" attacks, such as through
sparse attacks. For example these include small stickers on a road sign,  which
may tamper with autonomous vehicles \cite{eykholt2017robust}. The purpose of
this work is adversarial image generation for a broad class of (possibly
non-differentiable) dissimilarity metrics for both undefended and defended
networks. We do not make judgment regarding which metric is ``best''; instead we
are interested in an attack framework that works well for a broad class of
metrics.   

Adversarial attacks are often broadly categorized into one of two types:
white-box attacks, where the full structure of the neural network is provided to
the attacker, including gradient information, or black-box attacks, where the
attacker is only given the model decision. One of the first proposed adversarial
attacks is the Fast Gradient Signed Method (FGSM), which generates an
adversarial image with respect to the $\ell_\infty$ norm, along with its
iterative form, dubbed Iterative FGSM (IFGSM) \cite{FGSM,ifgsm}. A similar
iterative attack was also done with respect to the $\ell_2$ norm. In their
purest form, the above attacks perform gradient ascent on the training loss
function subject to a norm constraint on the perturbation, either with one step
in the case of FGSM, or multiple steps in the case of IFGSM, and their $\ell_2$
norm equivalents. Apart from training loss maximization, attacks have been
developed using loss functions that \emph{directly} measure misclassification
\cite{cw,deepfool}. Others have considered the $\ell_1$ and $\ell_0$ norms;
these both induce sparsity in the perturbations \cite{sparsefool}. In the
black-box setting, adversarial examples are generated using only model
decisions, which is a much more expensive
endeavor. However, black-box methods often perform better, most notably by avoiding gradient obfuscation, since they take advantage of sampling properties near the decision boundary of the model. Notable examples of black-box (decision-based) attacks are the Boundary Attack \cite{boundaryattack} and the recent HopSkipJumpAttack \cite{bapp}.

The development of new and improved adversarial attacks has occurred in parallel with various defensive training regimes to provide robustness against adversarial perturbations. The task of training a robust network is two-fold: models must be resistant to perturbations of a certain magnitude, while also maintaining classification ability on clean data. It has been argued that these two objectives are inherently ``at odds" \cite{atodds}. A popular method for training robust networks is \textit{adversarial training}, where adversarial examples are added to the training data (see for example \cite{madryLinf}). 
\subsection*{Contributions} This paper introduces an attack methodology for not
just $\ell_p$ norms, but any adversarial dissimilarity metric with a closed
proximal form. This includes, but is not limited to, $\ell_1$, $\ell_2$,
$\ell_\infty$, the $\ell_0$ counting ``norm", i.e. a true measurement of
sparseness of the perturbation, and total variation, a non-$\ell_p$
dissimilarity. Our approach adopts the relaxation structure of the recently
proposed LogBarrier attack \cite{logbarrier}, which required differentiable
metrics. We extend this work to include a broad class of non-smooth
(non-differentiable) metrics. Our algorithm, ProxLogBarrier, uses the proximal
gradient method for generating adversarial perturbations. We demonstrate our
attack on MNIST, CIFAR10, and ImageNet-1k datasets. ProxLogBarrier shows
significant improvement over both the LogBarrier attack, and over the other
attacks we considered. In particular, in the $\ell_0$ case, we achieve
state-of-the-art results with respect to a suite of attacks typically used for
this problem class. Finally, by using the total variation dissimiliarity, we shed
light on a new class of imperceptible adversaries that incorporates neighboring
pixel information, which can be viewed as an adversarial attack measured in a convolutional
norm.

\section{Background material}
\subsection{Adversarial attacks}
Let $\cX$ be the image space, and $\Delta_c$ be the label space (the
unit-simplex for $c$ classes). An image-label pair is defined by $(x,y) \in
\cX \times \Delta_c$, with the image belonging to one of $c$ classes. The
trained model is defined by $f : \cX \to \Delta_c$. An adversarial perturbation
should be small with respect to a \textit{dissimilarity   metric} (henceforth simply called the metric) $m(\cdot;x)$, e.g.  $\| \cdot -   \ x\|_\infty$. Formally, the optimal adversarial perturbation is the minimizer of the following optimization problem:
\begin{align}\label{eq: main}
    \min_{u \in \cX} \ m(u;x) \quad \text{subject to} \quad \text{argmax }f(u)\neq y.
\end{align}
DNNs might be powerful classifiers, but that does not mean their decision
boundaries are well-behaved. Instead, researchers have popularized using the
training loss, often the cross-entropy loss, as a surrogate for the decision
boundary: typically a model is trained until the loss is very low, which is
often related to good classification performance. Thus, instead of solving
\eqref{eq: main}, one can perform Projected Gradient Descent (PGD) on the cross-entropy loss:
\begin{equation}
    \max_{u \in \cX} \ \mathcal{L}(u) \quad \text{subject to} \quad  m(u;x) \leq \varepsilon,
\end{equation}
where $m(\cdot;x)$ is typically taken to be either the $\ell_2$ or $\ell_\infty$ norm, and $\varepsilon$ defines the perturbation threshold of interest. 

Some adversarial attack methods try to solve the problem posed in (\ref{eq:
main}) without incorporating the loss function used to train the network. For
example, Carlini \& Wagner attack the logit-layer of a network and solve a different optimization problem, which depends on the choice of norm \cite{cw}. Regarding adversarial defense methods, they demonstrated how a significant number of prior defense methods fail because of ``gradient obfuscation", where   gradients are small only locally to the image \cite{obfuscated_cw}. Another metric of adversarial dissimilarity is the $\ell_0$ ``norm", which counts the number of total different pixels between the adversary and the clean image \cite{sparsefool,jsma}. This is of interest because an adversary might be required to also budget the number of allowed pixels to perturb, while still remaining ``imperceptible" to the human eye. For example, the sticker-attack \cite{eykholt2017robust}  is a practical attack with real-world consequences, and does not interfere with every single part of the image.
\subsection{Proximal gradient method}\label{sec:prox}
Our adversarial attack amounts to a proximal gradient method. Proximal algorithms are a driving force for nonsmooth optimization problems, and are receiving more attention in the deep learning community on a myriad of problems \cite{proxquant,admm-prox,learningprox,catalyst}. For a full discussion on this topic, we suggest \cite{beckbook}.  

We consider the following framework for proximal algorithms, namely a composite minimization problem \begin{align}\label{eq:composite}
    \min_{x \in \cE } \Phi(x) := f(x) + g(x) 
\end{align}
where $\cE$ is a Euclidean space. We make the following assumptions:
\begin{itemize}
    \item $g$ is a non-degenerate, closed convex function over $\cE$ 
    \item $f$ is non-degenerate, closed function, with $\text{dom}(f)$ convex, and has $L$-Lipschitz gradients over the interior of its domain
    \item $\text{dom}(g) \subseteq \text{int}(\text{dom}(f))$
    \item the solution set, $S$, is non-empty.
\end{itemize}
Generating a stationary point of \eqref{eq:composite} amounts to finding a fixed point of the following sequence:
\begin{align}\label{eq:prox_iters}
    x^{(k+1)} = \text{Prox}_{\tau}g(x^{(k)} - \tau \nabla f (x^{(k)})),
\end{align}
where $\tau > 0$ is some step size, and $\text{Prox}_\lambda g(\cdot)$ is defined as $$ \text{Prox}_\lambda g(x) := \argmin_{u \in \mathcal{E}} g(u) + \dfrac{1}{2\lambda}\|u - x \|_2^2. $$
Despite $f$ not being convex, there are still convergence properties we can get from a sequence of iterates generated in this way. The following theorem is a simplified version of what can be found in \cite{beckbook} (Section 10.3 with proof), and is the main motivation for our proposed method.
\begin{theorem}
Given the assumptions on (\ref{eq:composite}), let $\{x^k\}_{k\geq0}$ be the sequence generated by (\ref{eq:prox_iters}), with fixed step size $\tau \in (\frac{L}{2},\infty)$. Then,
\begin{enumerate}[(a)]
    \item the sequence $\{\Phi(x^k\}_{k\geq0}$ is non-increasing. In addition, $\Phi(x^{k+1}) < \Phi(x^k)$ if and only if $x^k$ is not a stationary point of (\ref{eq:composite});
    \item $ \tau \left(x^k - \text{Prox}_{\frac{1}{\tau}}(x^k - \frac{1}{\tau} \nabla f(x^k)) \right) \to 0$ as $k \to \infty$;
    \item all limit points of the sequence $\{x^k\}_{k\geq0}$ are stationary points of (\ref{eq:composite}).
\end{enumerate}
\end{theorem}

\section{Our method: ProxLogBarrier}
Following the previous theoretical ideas, we reformulate (\ref{eq: main}) in the following way:
\begin{align}\label{eq: reform}
    \min_{u \in \cX} \ m(u;x) \quad \text{s.t.} \quad z_{\max} - z_y > 0.
\end{align}
Here, $Z(\cdot)$ is the model output before the softmax layer that ``projects" onto $\Delta_c$, and so $z_{\max} := \max_{i\neq y} [Z(u)]_i$ and $z_y := [Z(u)]_y $. In other words, we want to perturb the clean image minimally in such a way that the model misclassifies it. This problem is difficult as the decision boundary has virtually no exploitable structure. Thus the problem can be relaxed using a logarithmic barrier, a technique often used in traditional optimization \cite{nocedal}, 
\begin{align}\label{eq: logbar1}
    \min_{u \in \cX} \ m(u;x) - \lambda \log( z_{\max} - z_y).\end{align}
This objective function now includes the constraint that enforces misclassification. In \cite{logbarrier}, (\ref{eq: logbar1}) was originally solved via gradient descent, which necessarily assumes that $m(\cdot;x)$ is at least differentiable. The assumption of differentiability is not a given, and may be impracticable. For example, consider the subgradient of $\ell_\infty$ for an element in $\R^n$;
$$ \partial \| \cdot \|_\infty(x) = \text{sign}(x_k) e_k, $$ where $k := \argmax_i \{ |x_i| \}$, and $\{e_i\}_{i=1}^n$ are the standard basis vectors. At each subgradient step, very little information is obtained. Indeed, in the
original LogBarrier paper, a smooth approximation of this norm was used to get around this issue. We shall see that this does not occur with our proposed ProxLogBarrier method.

For brevity, let $\varphi(\cdot) := -\log(\cdot)$ and $$ F(u) := \left(\max_{i\neq y} Z(u)\right) - [Z(u)]_y. $$  The optimization problem \eqref{eq: logbar1} becomes
\begin{align}\label{eq: logbar}
    \min_{u \in \cX} \  m(u;x) +  \lambda \varphi(F(u)).
\end{align}
One can draw several similarities between (\ref{eq: logbar}) and (\ref{eq:composite}). As before, we have no guarantees of convexity on  $\varphi \circ F$, which is a representation of $f$ in the composite problem, but it is smooth provided  $F(\cdot) \in \text{dom}(\varphi)$ (that is, $F$ is smooth from a computational perspective). Our dissimilarity metric $m(u;x)$ represents $g$, as it usually has a closed-form proximal operator. Thus, we simply turn to the proximal gradient method to solve the minimization problem in (\ref{eq: logbar}). 

\begin{figure}
  \centering
  \includegraphics[width=2in]{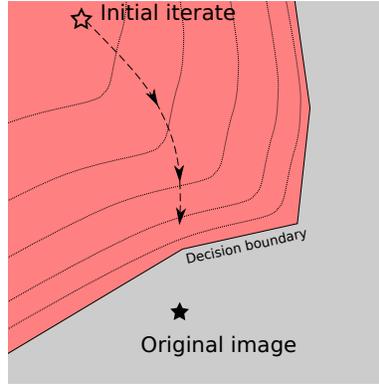}
  \caption{Illustration of the ProxLogBarrier attack: the attack is initialized with a
  misclassified image, which is then moved towards the original image.}
  \label{fig:illustration}
\end{figure}

We iteratively find a minimizer for the problem; the attack is outlined in Algorithm \ref{algo: plb-algo}. Due to the highly non-convex nature of the decision boundary, we perform a backtracking step to ensure the proposed iterate is in fact adversarial. We remark that the adversarial attack problem is constrained by the image-space, and thus requires a further  projection step back onto the image space (pixels must be in the range [0,1]). In traditional non-convex optimization, best practice is to also record the ``best iterate", as valleys are likely pervasive throughout the decision boundary. This way, even at some point our gradient sends our image far-off and is unable to return in the remaining iterations, we already have a better candidate.  The algorithm begins with a misclassified image, and moves the iterates towards the original image by minimizing the dissimilarity metric. Misclassification is maintained by the log barrier function, which prevents the iterates from crossing the decision boundary. Refer to Figure \ref{fig:illustration}. Contrast this with PGD based algorithms, which begin at or near the original image, and iterate away from the original image.
\begin{algorithm}
\caption{$\text{ProxLogBarrier (PLB)}$ }
\label{algo: plb-algo}
\begin{algorithmic}
\STATE Input: image-label pair $(x,y)$, trained model $f$, adversarial dissimilarity metric $m(\cdot;x)$
\STATE Intialize hyperparameters: $ K_{\text{inner}}, K \in \mathbb{N}$, and  $\mu, h,  \lambda_0 >0, \beta\in(0,1)$. 
\STATE Initialize $u^{(0)}$ to be misclassified, $w^{(0)} := u^{(0)}$ 
\FOR{$k = 0, 1, 2, \ldots, K$}
\STATE Every $K_{\text{inner}}$ iterations: $\lambda = \lambda_0\beta^k$
\STATE $y^{(k)} = \text{Prox}_{\mu m }\left(u^{(k)} - h \lambda  \nabla  \varphi(F(u^{(k)}))\right)$
\STATE $u^{(k+1)} = \text{Project}(y^{(k)};\mathcal{X})$
\STATE Backtrack along line between current and previous iterate until misclassified
\IF{$m(u^{(k+1)};x) < m(w^{(k)};x)$}
\STATE $w^{(k+1)} = u^{(k+1)}$
\ELSE \STATE $w^{(k+1)} = w^{(k)}$
\ENDIF
\ENDFOR
\STATE Output: $w^{(K)}$
\end{algorithmic}
\end{algorithm}

\subsection*{Proximal operators for $\ell_p$ dissimilarities}
To complete the algorithm, it remains to compute the proximal operator $\text{Prox}_{\mu m}(\cdot)$ 
for various choices of $m$. One can turn to \cite{beckbook} for complete derivations of the proximal
operators for the adversarial metrics we are considering, namely $\ell_1, \ell_2, \ell_\infty$ norms, 
and the $\ell_0$ cardinality function. Consider measuring the $\ell_\infty$ distance between the 
clean image and our desired adversarial perturbation: $$ m(u;x) := \|u - x\|_\infty.$$Due to the 
Moreau Decomposition Theorem \cite{RWbible}, the proximal operator of this function relies on 
projecting onto the unit $\ell_1$ ball:
\begin{align}
    \text{Prox}_{\mu\|\cdot \ - \ x\|_\infty}(z) &= x + \text{Prox}_{\mu\| \cdot \|_\infty}(z - x) \nonumber \\ 
    &= x + (z - x) - \mu \text{Prox}_{\mathbb{B}_1}((z - x)/\mu) \nonumber \\
    &= z - \mu \text{Proj}_{\|\cdot\|_1}((z - x)/\mu) \nonumber.
\end{align}
We make use of the algorithm from \cite{duchi} to perform the projection step, implemented over batches of vectors for efficiency. Similarly, one obtains the proximal operator for $\ell_1$ and $\ell_2$ via the same theorem, 
\begin{align}
    & \text{Prox}_{\mu\|\cdot \ - \ x\|_1}(z) = x + \mathcal{T}_\mu(z - x), \nonumber \\
    & \text{Prox}_{\mu\|\cdot \ - \ x\|_2}(z) = z - \mu \text{Proj}_{\|\cdot \|_2}((z - x)/\mu), \nonumber
\end{align}
where $\mathcal{T}_\mu(s) := \text{sign}(s)\max\{|s|-\mu,0\}$ is the soft thresholding operator. In the case that one wants to minimize the number of perturbed pixels in the adversarial image, one can turn to the counting ``norm", called $\ell_0$, which counts the number of non-zero entries in a vector. While this function is non-convex, the proximal operator still has a closed form:
\begin{align}
     P_{\mu\|\cdot \  - \  x\|_0}(z) &= x + \mathcal{H}_{\sqrt{2\mu}}(z - x) \nonumber
\end{align}
where $\mathcal{H}_\alpha(s) = s\bm{1}_{\{|s| > \alpha \}}(s)$ is a hard-thresholding operator, and acts component-wise in the case of vector arguments.
\subsection*{Example of non-$\ell_p$ dissimilarity: Total variation}
We let $\cX$ denote the image space, and for the time being assume the images are grayscale, and let $M$ denote the finite-difference operator on the grid-space defined by the image.  Then $M : \cX \to \cX \times \cX$, where
\begin{align}
    (Mv)_{i,j} = \begin{pmatrix}
          D_x v \\
          D_y v
          \end{pmatrix}_{i,j}
          := \begin{pmatrix}
          v_{i+1,j} - v_{i,j} \\
          v_{i,j+1} - v_{i,j}
          \end{pmatrix}_{i,j},
\end{align}
where $(i,j)$ are the pixel indices of the image in row-column notation. The \textit{anisotropic total variation semi-norm} is defined by 
\begin{align}
    \| v \|_{\text{TV}} := \| Mv \|_{1,1} = \sum_{i,j} | (D_x v)_{i,j} | + |(D_y v)_{i,j}|,
\end{align}
where  $\|\cdot\|_{1,1}$ is an induced matrix norm. Heuristically, this is a
measure of large changes between neighboring pixels. In practice $Mv$ can be
implemented via a convolution. In the case of color images,
we aggregate the total variation for each channel. Total variation (TV) is not
true norm, in that non-zero images $v$ can have zero TV. In what
follows, we omit the distinction and write TV-norm to mean the
total variation seminorm. Traditionally, TV has been used in the context of image denoising
\cite{ROF}.

What does this mean in the context of adversarial perturbations? The TV-norm of
the perturbation
will be small when the perturbation has few jumps between pixels. That is, small
TV-norm perturbations have \textit{locally flat regions}. This is primarily
because TV-norm is \textit{convolutional} in nature: the finite-difference
gradient operator incorporates neighboring pixel information. We note that this
is not the first instance of TV being used as a disimillarity metric
\cite{spatially}; however our approach is quite different and is not derived
from a flow. An outline for the proximal operator can be found in
\cite{beckbook}; we use a standard package for efficient computation
\cite{proxtv_1,proxtv_2}
\begin{table*}[b]
  \caption{Adversarial robustness statistics, measured in the $\ell_0$ norm.}
\label{tab:l0-stats}
  \begin{center}
\begin{adjustbox}{width=\textwidth}
  \begin{tabular}{lc rrc c rrrc rrrc}
  \toprule
  & & \multicolumn{3}{c}{MNIST} & & \multicolumn{3}{c}{CIFAR10} & & \multicolumn{3}{c}{ImageNet} \\ 
  & & \multicolumn{2}{c}{\% error at}  &  \multirowcell{2}{median\\distance} & &
  \multicolumn{2}{c}{\% error at}  & \multirowcell{2}{median\\distance} & &
  \multicolumn{2}{c}{\% error at}  & \multirowcell{2}{median\\distance}  \\ 
  \cmidrule(lr){3-4} \cmidrule(lr){7-8} \cmidrule(lr){11-12}
  & &$\varepsilon=10$ & $\varepsilon=30$ & & &$\varepsilon=30$ & $\varepsilon=80$ & & &$\varepsilon=500$ & $\varepsilon=1000$ &  \\
  \midrule
  $\text{PLB}$   &  & \textbf{86.30} & \textbf{100} &  \textbf{6 }  && \textbf{44.10} &\textbf{ 68.50} & \textbf{39}   && \textbf{66.00} & \textbf{80.20}  & \textbf{268}    \\
  SparseFool   &  & 46.00 & 99.40 &  11    && 15.60  &  22.60 &  3071\footnotemark && 30.40 & 46.80  & $1175^*$   \\
  JSMA         &  & 12.73 & 61.38 &   25  && 29.56  & 48.92 & 84   && ---  & --- & ---   \\
  Pointwise    &  & 5.00 & 57.30 & 28   && 13.20 & 50.60  & 80   && --- & --- &   --- \\
  \toprule
   (D) $\text{PLB}$   &  &  \textbf{79.8} &  \textbf{98.90} &   \textbf{6}   && \textbf{74.90} &  \textbf{97.80} &   \textbf{13}    && \textbf{38.40} & \textbf{70.0}   & \textbf{691} \\
   (D) SparseFool   &  & 20.67 &  75.45 &  20 &&   34.23 &  52.15 &   70  && 24.80 & 41.80  & $1310^*$   \\
   (D) JSMA         &  & 12.63 &  44.51 &  34  &&   36.65 &  60.79 &   53    && ---  & --- & ---   \\
   (D) Pointwise    &  & 12.50 &  65.80 &  24 &&  23.80 &  43.10 &  102    && --- & --- &   --- \\
\bottomrule
\end{tabular}
\end{adjustbox}
\end{center}
\end{table*}

\begin{table*}[t]
  \caption{Adversarial robustness statistics, measured in the $\ell_\infty$ norm.}
\label{tab:linf-stats}
  \begin{center}
\begin{adjustbox}{width=\textwidth}
  \begin{tabular}{lc rrrc c rrrc rrrc}
  \toprule
  & & \multicolumn{3}{c}{MNIST} & & \multicolumn{3}{c}{CIFAR10} & & \multicolumn{3}{c}{ImageNet} \\ 
  & & \multicolumn{2}{c}{\% error at}  &  \multirowcell{2}{median\\distance} & &
  \multicolumn{2}{c}{\% error at}  & \multirowcell{2}{median\\distance} & &
  \multicolumn{2}{c}{\% error at}  & \multirowcell{2}{median\\distance}  \\ 
  \cmidrule(lr){3-4} \cmidrule(lr){7-8} \cmidrule(lr){11-12}
  & &$\varepsilon=0.1$ & $\varepsilon=0.3$ & & &$\varepsilon=\frac{2}{255}$ & $\varepsilon=\frac{8}{255}$ & & &$\varepsilon=\frac{2}{255}$ & $\varepsilon=\frac{8}{255}$ &  \\
  \midrule
  $\text{PLB}$   & & 10.30 &  \textbf{100} & \textbf{ \num{1.67e-1}} &&  \textbf{95.00} &  \textbf{98.60} &  \textbf{\num{2.88e-3}} && 20.40& 33.80 &   \num{6.66e-2} \\
  PGD         & & \textbf{10.70} &   80.90& \num{1.76e-1} && 54.70 &  87.00 &  \num{5.91e-3} && 90.80 & 98.60 &  \textbf{ \num{2.5e-3}} \\
  DeepFool     & &  8.12 &   86.55 &  \num{2.25e-1}   && 16.23 & 51.00   &  \num{3.04e-2}  && \textbf{93.64} & \textbf{100}  &   {\num{2.8e-3}} \\
  LogBarrier         & & 5.89  & 73.90  & \num{2.43e-1} && 60.60 &  93.10  & \num{6.84e-3} && 7.60 & 7.70 &  \num{6.16e-1}  \\
  \toprule
   (D) $\text{PLB}$   & &   \textbf{3.0} & \textbf{ 32.9} &  \textbf{\num{3.24e-1}} && 23.3 &  44.1 &  \num{3.64e-2}  && 11.40  & 18.80  & \num{1.06e-1} \\
   (D) PGD         & &  2.8 &  23.6 &  \num{3.37e-1} && 22.9 &  \textbf{46.1} & \textbf{ \num{3.45e-2}}   && \textbf{49.20}  & 96.60 & \textbf{\num{7.94e-3}} \\
   (D) DeepFool     & &   2.7 &  10.2 &  \num{6.66e-1}    && \textbf{ 23.8} &  44.1 &  \num{3.74e-2}   && 43.20  & \textbf{97.40}  & \num{9.31e-3} \\
   (D) LogBarrier         & & 2.50 & 11.89  & \num{5.48e-1} && 17.6 &  28.3 &  \num{8.01e-2} && 9.80 & 10.40 &  \num{4.43e-1}  \\
  \bottomrule
\end{tabular}
\end{adjustbox}
\end{center}
\end{table*}


\section{Experimental methodology}
\subsection*{Outline}
We compare the ProxLogBarrier attack with several other adversarial attacks on MNIST \cite{mnist_dataset}, CIFAR10 \cite{cifar10_dataset}, and ImageNet-1k \cite{imagenet_dataset}. For MNIST, we use the network described in \cite{jsma}; on CIFAR10, we use a ResNeXt network \cite{resnext}; and for ImageNet-1k,  ResNet50 \cite{resnet50,DAWNBench}. We also consider defended models for the aforementioned networks. This is to further benchmark the attack capability of the ProxLogBarrier, and to reaffirm previous work in the area. For defended models, we consider  Madry-style adversarial training for CIFAR10 and MNIST \cite{madryLinf}. On ImageNet-1k, we use the recently proposed scaleable input gradient regularization for adversarial robustness \cite{finlay2019scaleable}. We randomly select 1000 (test) images to evaluate performance on MNIST and CIFAR10, and 500 (test) images on ImageNet-1k.  We consider the same images on their defended counterparts. We note that for ImageNet-1k, we consider the problem of Top5 misclassification, where the log barrier is with respect to the following constraint set
$$  Z[u]_{(5)} - Z[u]_{(y)} > 0  $$ where $(i)$ denotes the $i^{\text{th}}$ largest index.
 
We compare the ProxLogBarrier attack with a wide range of attack algorithms that
are available through the FoolBox adversarial attack library \cite{foolbox}.
For perturbations in $\ell_0$, we compare against SparseFool \cite{sparsefool},
Jacobian Saliency Map Attack (JSMA) \cite{jsma}, and Pointwise
\cite{pointwise} (this latter attack is black-box). For $\ell_2$ attacks, we
consider Carlini-Wagner's attack (CW) \cite{cw}, Projected Gradient Descent (PGD) \cite{ifgsm}, DeepFool \cite{deepfool}, and the original LogBarrier attack \cite{logbarrier}. Finally, for $\ell_\infty$ norm perturbations, we consider PGD, DeepFool, and LogBarrier. All hyperparameters are left to their implementation defaults, with the exception of SparseFool, where we used the exact parameters indicated in the paper. We omit the One-Pixel attack \cite{onepixel}, as \cite{sparsefool} showed that this attack is quite weak on MNIST, CIFAR10, and not tractable on ImageNet-1k.
\begin{table*}[b]
  \caption{Adversarial robustness statistics, measured in the $\ell_2$ norm.}
\label{tab:l2-stats}
  \begin{center}
\begin{adjustbox}{width=\textwidth}
  \begin{tabular}{lc rrrc c rrrc rrrc}
  \toprule
  & & \multicolumn{3}{c}{MNIST} & & \multicolumn{3}{c}{CIFAR10} & & \multicolumn{3}{c}{ImageNet} \\ 
  & & \multicolumn{2}{c}{\% error at}  &  \multirowcell{2}{median\\distance} & &
  \multicolumn{2}{c}{\% error at}  & \multirowcell{2}{median\\distance} & &
  \multicolumn{2}{c}{\% error at}  & \multirowcell{2}{median\\distance}  \\ 
  \cmidrule(lr){3-4} \cmidrule(lr){7-8} \cmidrule(lr){11-12}
  & &$\varepsilon=1.25$ & $\varepsilon=2.3$ & & &$\varepsilon=\frac{80}{255}$ & $\varepsilon=\frac{120}{255}$ & & &$\varepsilon=0.5$ & $\varepsilon=1$ &  \\
  \midrule
  $\text{PLB}$   & &\textbf{ 38.60} & \textbf{99.40}  & \textbf{1.35 } && \textbf{97.70} & \textbf{99.80}  &  \textbf{\num{1.15e-1}} && \textbf{47.60 }& \textbf{89.40} &\textbf{ \num{5.24e-1}} \\
  CW          & & 35.10  & 98.30 & 1.41   && 89.94 & 95.97 &   \num{1.32e-1} && 20.06 &  44.26 &  1.16   \\
  PGD        & & 24.70 & 70.00 & 1.70   && 60.60 & 73.30  &   \num{2.10e-1} &&  37.60 &  70.60 & \num{6.72e-1} \\
  DeepFool     & & 13.21 & 48.04  &  2.35 && 17.33  & 22.04 & 1.11  &&  40.08 &  76.48 &  \num{6.23e-1} \\
  LogBarrier     & & 37.40 & 98.90  & 1.35  && 69.60  & 84.00 & \num{2.02e-1}  &&  43.70 &  88.30 &  \num{5.68e-1} \\
  \toprule
   (D) $\text{PLB}$   & & \textbf{29.50} &  \textbf{92.90} &  \textbf{1.54} && 28.7 &  35.4 &  \num{7.26e-1}  && \textbf{15.80} & \textbf{28.20} &\textbf{ 1.74} \\
   (D) CW         & & 28.24 &  78.59  &  1.72  &&  \textbf{29.6} &  \textbf{38.7 } &  \textbf{\num{6.60e-1}}  && --- & --- & --- \\
   (D) PGD         & & 17.20 &  45.70 &  2.44 && 28.30 &  34.70 &  \num{7.97e-1}  && 14.60 & 22.60 & 2.20 \\
   (D) DeepFool     & &   5.22 &  18.07 &  3.73    &&  28.0 &  33.3 &  \num{9.31e-1}   && 15.60 & 24.40   & 2.14   \\
   (D) LogBarrier         & & 25.00 & 89.60  & 1.65  && 28.0 &  34.6 &  \num{7.36e-1}  && 10.00 & 10.20 & 63.17 \\
  
  \bottomrule
\end{tabular}
\end{adjustbox}
\end{center}
\end{table*}
\footnotetext{We believe this is an implementation error on behalf of the repository. To accurately compare, we attacked an 18-layer ResNet for CIFAR10 that achieves slightly worse clean error as reported in \cite{sparsefool}. Our median percent pixels perturbed was 1.4\%, and they reported 1.27\%. }
\subsection*{Implementation details for our algorithm}
When optimizing for $\ell_2$ based noise, we initialize the adversarial image with sufficiently large Gaussian noise; for $\ell_\infty$ and $\ell_0$ based perturbations, we use uniform noise. For hyper-parameters, we used $\lambda_0 = 0.1, \beta = 0.75, h = 0.1, \mu = 1$, with $K = 900, K_{\text{inner}}=30$. We observed some computational drawbacks for ImageNet-1k: firstly, the proximal operator for the $\ell_0$ norm is far too strict. We decided to use the $\ell_1$ norm to induce sparseness in our adversarial perturbation (changing both the prox parameter and the step size to $0.5$). Other parameter changes for the ImageNet-1k dataset are that for the proximal parameter in the $\ell_\infty$ case, we set $\mu=3$, and we used 2500 algorithm iterations. Finally, we found that using the softmax layer outputs helps with ImageNet-1k attacks against both the defended and undefended network. For TV-norm, perturbations, we set the proximal parameter $\mu=5$, and $K=200$ with $K_{\text{inner}}=20$ (far less than before). 

\subsection*{Reporting}
For perturbations in $\ell_2$ and $\ell_\infty$, we report the percent misclassification at various threshold levels that are somewhat standard \cite{atodds}. Our choices for $\ell_0$ distance thresholds were arbitrary, however we supplement with  a median perturbation distances on all attack norms to mitigate cherry-picking. For attacks that were unable to successfully perturb at least half the sampled images, we do not report anything. If the attack was able to perturb more than half but not all, we add an asterisk to the median distance. We denote the defended models by ``(D)" (recall that for MNIST and CIFAR10, we are using Madry's adversarial training, and scaleable input-gradient regularization for Imagenet-1k).

\subsubsection*{Perturbations in \boldmath{$\ell_0$}}
\begin{figure}[t]
    \centering
    \begin{subfigure}[b]{0.25\textwidth}
        \includegraphics[width=\textwidth]{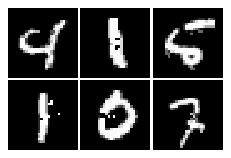}
        \caption{$\ell_0$ attacks on MNIST}
        \label{fig:mnist_l0}
    \end{subfigure}
    \hspace{2em} 
    \begin{subfigure}[b]{0.25\textwidth}
        \includegraphics[width=\textwidth]{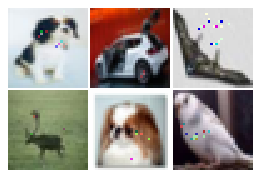}
        \caption{$\ell_0$ attacks on CIFAR10}
        \label{fig:cifar10_l0}
    \end{subfigure}
    \caption{Adversarial images for  $\ell_0$ perturbations, generated by
      our method. 
  }\label{fig:pics}
\end{figure}
Result for $\ell_0$ perturbations are found in Table \ref{tab:l0-stats}, with examples available in Figure \ref{fig:pics} and Figure \ref{fig:imagenet_l0}. Across all datasets considered, ProxLogBarrier outperforms all other attack methods, for both defended and undefended networks. It also appears immune to Madry-style adversarial training on both MNIST and CIFAR10. This is entirely reasonable, for the Madry-style adversarial training is targeted towards $\ell_\infty$ attacks. In contrast, on ImageNet-1k, the defended model trained with input-gradient regularization performs significantly better than the undefended model, even though this defence is not aimed towards $\ell_0$ attacks. Neither JSMA or Pointwise scale to networks on ImageNet-1k. Pointwise exceeds at smaller images, since it takes less than 1000 iterations to cycle over every pixel and check if it can be zero'd out. We remark that SparseFool was unable to adversarially attack all images, whereas ProxLogBarrier always succeeded.

\subsubsection*{Perturbations in \boldmath{$\ell_\infty$}}
Results for $\ell_\infty$ perturbations are found in Table \ref{tab:linf-stats}. Our attack stands out on MNIST, in both the defended and undefended case. On CIFAR10, our attack is best on the undefended network, and only slightly worse than PGD when adversarially defended. On ImageNet-1k, our method suffers dramatically. This is likely due to very poor decision boundaries with respect to this norm $\ell_\infty$, as our method will necessarily be better when the boundaries are not muddled. PGD does not focus on the decision boundaries explicitly, thus has more room to find something adversarial quickly. 
\begin{figure}
    \centering
    \begin{subfigure}[b]{0.15\textwidth}
        \includegraphics[width=\textwidth]{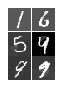}
        \caption{TV-norm attacks on MNIST}
        \label{fig:mnist_tv}
    \end{subfigure}
    \hspace{2em} 
    \begin{subfigure}[b]{0.15\textwidth}
        \includegraphics[width=\textwidth]{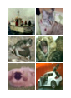}
        \caption{TV-norm attacks on CIFAR10}
        \label{fig:cifar10_tv}
    \end{subfigure}
    \caption{Adversarial images for  TV-norm perturbations, generated by
      our method. 
  }\label{fig:pics_tv}
\end{figure}
\subsubsection*{Perturbations in \boldmath{$\ell_2$}}
Results for perturbations measured in Euclidean distance are found in Table  \ref{tab:l2-stats}. For MNIST and ImageNet-1k, on both defended and undefended networks, our attack performs better than all other methods, both in median distance and at a given perturbation norm threshold. On CIFAR10, we are best on undefended but lose to CW in the defended case. However, the CW attack did not scale to ImageNet-1k using the implementation in the FoolBox attack library. 

\subsubsection*{Perturbations in the {TV-norm}}
\begin{figure}[t!]
    \centering
    \begin{subfigure}[b]{0.5\textwidth}
        \includegraphics[width=\textwidth]{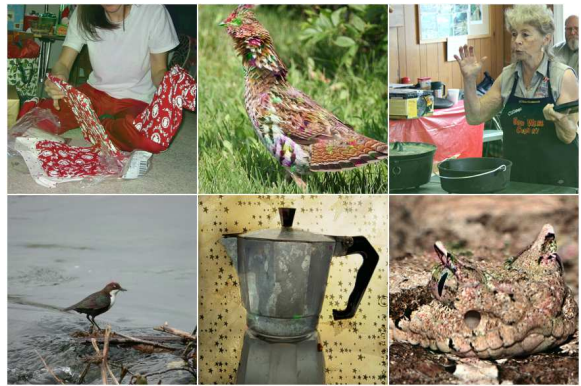}
        \caption{TV-norm attacks}
        \label{fig:imagenet_tv}
    \end{subfigure}
    \hspace{2em} 
    \begin{subfigure}[b]{0.5\textwidth}
        \includegraphics[width=\textwidth]{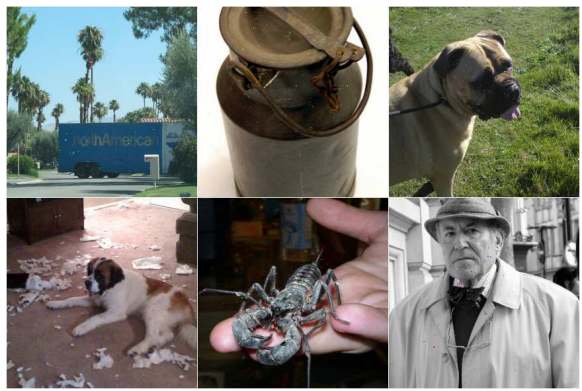}
        \caption{$\ell_0$ attacks, with fewer than 1000 pixels perturbed}
        \label{fig:imagenet_l0}
    \end{subfigure}
    \caption{Adversarial images for ImageNet-1k. Note that $\ell_0$ attacks are
      only visible when the image is magnified. The TV-norm perturbations are
      visible as either a tint of the full image, or as a set of local tints.
  }\label{fig:pics_imagenet}
\end{figure}

To our knowledge, there are no other TV-norm atacks against which to compare our
methods. However, we present the median total variation across the data in
question, and a handful of pictures for illustration. On MNIST, adversarial
images with minimal total variation are often as expected: near-flat
perturbations or very few pixels perturbed (see Figure \ref{fig:mnist_tv}). For CIFAR10 and ImageNet-1k, we have
found that adversarial images with
small TV-norm have an adversarial ``tint" on the image: they appear nearly
identical to the original, with a small color shift. When the
adversary is not a tint, perturbations are highly localized or localized in
several regions. See for example Figures \ref{fig:cifar10_tv} and \ref{fig:imagenet_tv}.
\begin{table}
  \caption{Statistics for perturbations in TV-norm}
\label{tab:runtimes}
  \begin{center}
\begin{adjustbox}{width=0.35\textwidth}
  \begin{tabular}{l ccc}
    \toprule
   & &  median TV-norm & max TV-norm\\
   \midrule
   MNIST      &  & 2.52 & 11.0 \\
   CIFAR10    &  & 1.36 & 11.0 \\
   ImageNet-1k &  & 13.4 & 149.6 \\
  \bottomrule
  \end{tabular}
\end{adjustbox}
\end{center}
\end{table}
\subsubsection*{Algorithm runtime}
We strove to implement ProxLogBarrier so that it could be run in a reasonable amount of time. For that reason, ProxLogBarrier was implemented to work over a batch of images. Using one consumer grade GPU, we can comfortably attack several MNIST and CIFAR10 images simultaneously, but only one ImageNet-1k image at a given time. We report our algorithm runtimes in Table \ref{tab:runtimes}. Algorithms implemented from the FoolBox repository were not written to take advantage of the GPU, hence we omit run-time comparisons. Heuristically speaking, PGD is one of the faster algorithms, whereas CW, SparseFool, and DeepFool are slower. We omit the computational complexity for minimizing total variation since the proximal operator is coded in C, and not Python.

We are not surprised that our attack in $\ell_0$ takes longer than the other norms; this is likely due to the backtracking step to ensure misclassification of the iterate. On ImageNet-1k, the ProxLogBarrier  attack in the $\ell_\infty$ metric is quite slow due to the projection step onto the $\ell_1$ ball, which is $\mathcal{O}(n\log(n))$, where $n$ is the input dimension size \cite{duchi}. 
\begin{table}
  \caption{ProxLogBarrier attack runtimes (in seconds)}
\label{tab:runtimes}
  \begin{center}
\begin{adjustbox}{width=0.45\textwidth}
  \begin{tabular}{l cccc}
    \toprule
   & Batch Size& $\ell_0$ & $\ell_2$ & $\ell_\infty$\\
   \midrule
   MNIST & 100 & 8.35 & 6.91&6.05 \\
   CIFAR10  & 25 & 69.07 &56.11 &30.87 \\
  ImageNet-1k & 1 & 35.45&29.47 & 75.50 \\
  \bottomrule
  \end{tabular}
\end{adjustbox}
\end{center}
\end{table}

\section{Conclusion}
We have presented a concise framework for generating adversarial perturbations
by incorporating the proximal gradient method. We have expanded upon the
LogBarrier attack, which was originally only effective in $\ell_2$ and
$\ell_\infty$ norms, by addressing the $\ell_0$ norm case and the total
variation seminorm. Thus we have proposed a method unifying all three common
perturbation scenarios. Our approach requires fewer hyperparameter tweaks than
LogBarrier, and performs significantly better than many attack methods we
compared against, both on defended and undefended models, and across all norm
choices. We highlight that our method is, to our knowledge, the best choice for
perturbations measured in $\ell_0$, compared to all other methods available in
FoolBox. We also perform better than all other attacks considered on the MNIST
network with in the median distance and in commonly reported thresholds. 
The proximal gradient method points towards new forms of adversarial attacks,
such as those measured in the TV-norm, provided the attack's dissimilarity
metric has a closed proximal form.

{\small
\bibliographystyle{plain}
\bibliography{ref}
}

\end{document}